\documentclass{article} 
\pdfoutput=1
\usepackage{hyperref}
\usepackage{L,times}
\usepackage{url}
\usepackage{graphicx}
\usepackage{subfigure}
\usepackage{amsmath,amsfonts,amssymb}
 \usepackage{verbatim}
 \usepackage{footnote}
 \usepackage{tablefootnote}
\DeclareGraphicsExtensions{.eps,.ps,.jpg,.png}

\title{Learning Document Embeddings by \mbox{Predicting} N-grams for Sentiment Classification of Long Movie Reviews}

\author{Bofang Li , Tao Liu \& Xiaoyong Du\\
Department of Computer Science\\
Renmin University of China\\
Beijing, P.R. China \\
\texttt{\{libofang, tliu, duyong\}@ruc.edu.cn} \\
\And
Deyuan Zhang \\
School of Computer \\
Shenyang Aerospace University \\
Shenyang, Liaoning, P.R. China \\
\texttt{dyzhang@sau.edu.cn} \\
\AND
Zhe Zhao \\
Department of Computer Science\\
Renmin University of China\\
\texttt{helloworld@ruc.edu.cn}
}

\begin{document}

\maketitle

\begin{abstract}
Despite the loss of semantic information, bag-of-ngram based methods still achieve state-of-the-art results for tasks such as sentiment classification of long movie reviews. Many document embeddings methods have been proposed to capture semantics, but they still can't outperform bag-of-ngram based methods on this task. In this paper, we modify the architecture of the recently proposed Paragraph Vector, allowing it to learn document vectors by predicting not only words, but n-gram features as well. Our model is able to capture both semantics and word order in documents while keeping the expressive power of learned vectors. Experimental results on IMDB movie review dataset shows that our model outperforms previous deep learning models and bag-of-ngram based models due to the above advantages. More robust results are also obtained when our model is combined with other models. The source code of our model will be also published together with this paper.
\end{abstract}

\section{Introduction}
\label{section-1}
Sentiment analysis is one of the most useful and well-studied task in natural language processing. For example, the aim of movie review sentiment analysis is to determine the sentiment polarity of a review that an audience posted, which can be used in applications such as automatically movie rating. This type of sentiment analysis can often be considered as a classification task. Normally, training and test documents are first represented as vectors. A classifier is trained using training document vectors and their sentiment labels. Test document labels can be predicted using test document vectors and this classifier.

The quality of document vectors will directly affect the performance of sentiment analysis tasks. Bag-of-words or bag-of-ngram based methods have been widely used to represent documents. However, in these methods, each word or n-gram is taken as a unique symbol, which is different to other words or n-grams absolutely, and semantic information is lost.

For modeling semantics of words, word embeddings \citep{williams1986learning, bengio2003neural} is proposed, which has been successfully applied to many tasks such as chunking, tagging \citep{collobert2008unified, collobert2011natural}, parsing \citep{Socher2011ICML} and speech recognition \citep{schwenk2007continuous}. Following the success of word embeddings, sentence and document embeddings have been proposed for sentiment analysis. For sentence level sentiment analysis, models like recurrent neural network \citep{socher2013recursive}, convolutional neural network \citep{kalchbrenner2014convolutional, Kim2014EMNLP}, and skip thought vectors \citep{kiros2015skip} all achieved state-of-the-art results. But for document level sentiment analysis, different document embeddings models like convolutional neural network, weighted concatenation of word vectors\citep{Maas2011ACL}, recurrent neural network \citep{mikolov2012statistical}, deep Boltzmann machine \citep{Srivastava2013UAI}, and deep averaging network \citep{Iyyer2015ACL} still can't outperform bag-of-ngram based models such as NBSVM \citep{Wang2012ACL}.
Thus, more powerful document embeddings learning methods are needed for sentiment analysis.

\begin{figure}[h]
\begin{center}
\includegraphics[width=1.0\linewidth]{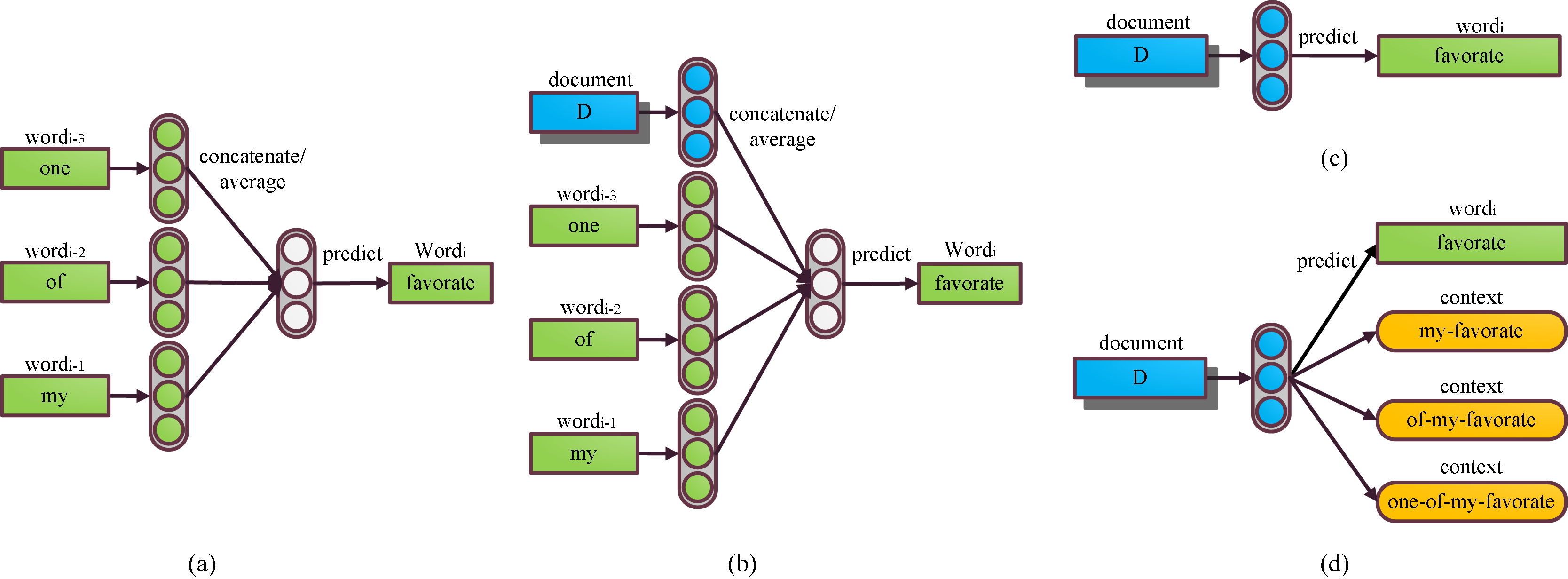}
\end{center}
\label{figure-abcd}
\vspace{-1.5em}
\caption{(a) CBOW. (b) PV. (c) simplified version of PV. (d) DV-ngram.}
\end{figure}

Recently, \citet{Le2014ICML} proposed a model of learning distributed representation for both sentences and documents, named as Paragraph Vector (PV). PV represents pieces of texts as compact low dimension continuous-value vectors. The process of learning PV is shown in Figure~\ref{figure-abcd}-b, which is similar with the typical word embeddings learning methods such as CBOW \citep{Mikolov2013NIPS} shown in Figure~\ref{figure-abcd}-a. PV basically treat each document as a special word and learn both document vectors and word vectors simultaneously by predicting the target word.

Vectors learned by PV are not sufficient for modeling documents. For example, when the learned information of word vectors of ``one'', ``of'', ``my'' is already sufficient for predicting the next word ``favorite'' (when the model in Figure~\ref{figure-abcd}-a is able to perform the prediction well enough), the document vector can't be sufficiently learned by the model of Figure~\ref{figure-abcd}-b. That is, the document vector predicts the word with the help of context, so it do not have to contains all the information. The expressive power of document vectors may be lost in this condition.

Due to this reason, we discover that a simplified version of PV shown in Figure~\ref{figure-abcd}-c is more effective for learning document vectors than PV in Figure~\ref{figure-abcd}-b
\footnote{ In contrast to our experimental results, \citet{Le2014ICML} reported that the simplified PV (referred to as PV-DBOW in their paper) is consistently worse than PV (referred to as PV-DM). But as pointed out by \citet{mesnil2014ensemble}, results reported by \citet{Le2014ICML} can only be reproduced when the data is not shuffled, which are considered invalid.}.
This simplified version of PV learns document vectors alone by predicting its belonging words, thus all the information can only be learned by document vectors to keep the expressive power. But this simplified version of PV does not take contextual words into consideration and thus word order information is lost. \footnote{As shown in our experiment section, simplified version of PV (DV-uni) outperforms PV 0.87 percent in terms of accuracy on IMDB dataset.}

In order to preserve the word order information, our model learns document vectors by predicting not only its belonging words, but n-gram features as well, as shown in Figure~\ref{figure-abcd}-d. Note that PV in figure~\ref{figure-abcd}-b may not be able to use n-gram features since there are no n-grams that can be specified given certain context. Similar to Paragraph Vector, we name our model as Document Vector by predicting ngrams (DV-ngram). More powerful document vectors can be learned using this model.

\section{Model}
\subsection{Basic Model for Modeling Semantics}
Traditional bag-of-words methods use one-hot representation for documents. Each word is taken as a unique symbol and is different to other words absolutely. This representation often ignores the impact of similar words to documents. For example, the distances among the first three documents in Table~\ref{table-d1234} are same in one-hot vector space, since there is only one different word. But from semantic point of view, $D_{1}$ is more similar to $D_{2}$ than to $D_{3}$. In order to solve this problem, the semantics of documents should be modeled. Distributed representation is a quite effective method for addressing this problem.

\begin{table}[t]
\caption{Illustration of documents for comparing document distance}
\label{table-d1234}
\begin{center}
\begin{tabular}{c l}
\hline
$D_{1}$  & I saw Captain American yesterday with my friends, its’ \underline{awesome}. \\
$D_{2}$  & I saw Captain American yesterday with my friends, its’ \underline{inspiring}. \\
$D_{3}$  & I saw Captain American yesterday with my friends, its’ \underline{meaningless}. \\
$D_{4}$  & I saw Captain American yesterday with my friends, its’ \underline{awesome and inspiring}. \\
\end{tabular}
\end{center}
\end{table}

Specifically, documents are represented by compact low dimension continuous-value vectors with randomly initialized values. Document vectors are learned by predicting which words belonging to them and which are not. Semantics such as synonyms can be modeled by document embeddings. For example, $D_{1}$ tends to be closer to $D_{4}$ in the new vector space, since they both need to predict the same word ‘awesome’. $D_{2}$ tends to be closer to $D_{4}$ due to the same reason. This will make $D_{1}$ to be much closer to $D_{2}$ than to $D_{3}$ since both $D_{1}$ and $D_{2}$ have the same neighbor $D_{4}$.

More formally, the objective of the document embeddings model is to maximize the following log probability
\begin{equation}\label{math-pwd}
  \sum_{i} \sum_{j} \log p\left( w_{i,j} | d_{i} \right)
\end{equation}
where $d_{i}$ denotes the $i^{th}$ document from document set $D$ and $w_{i,j}$ represents the $j^{th}$ word of $d_{i}$.

\begin{figure}[h]
\begin{center}
\includegraphics[width=1.0\linewidth]{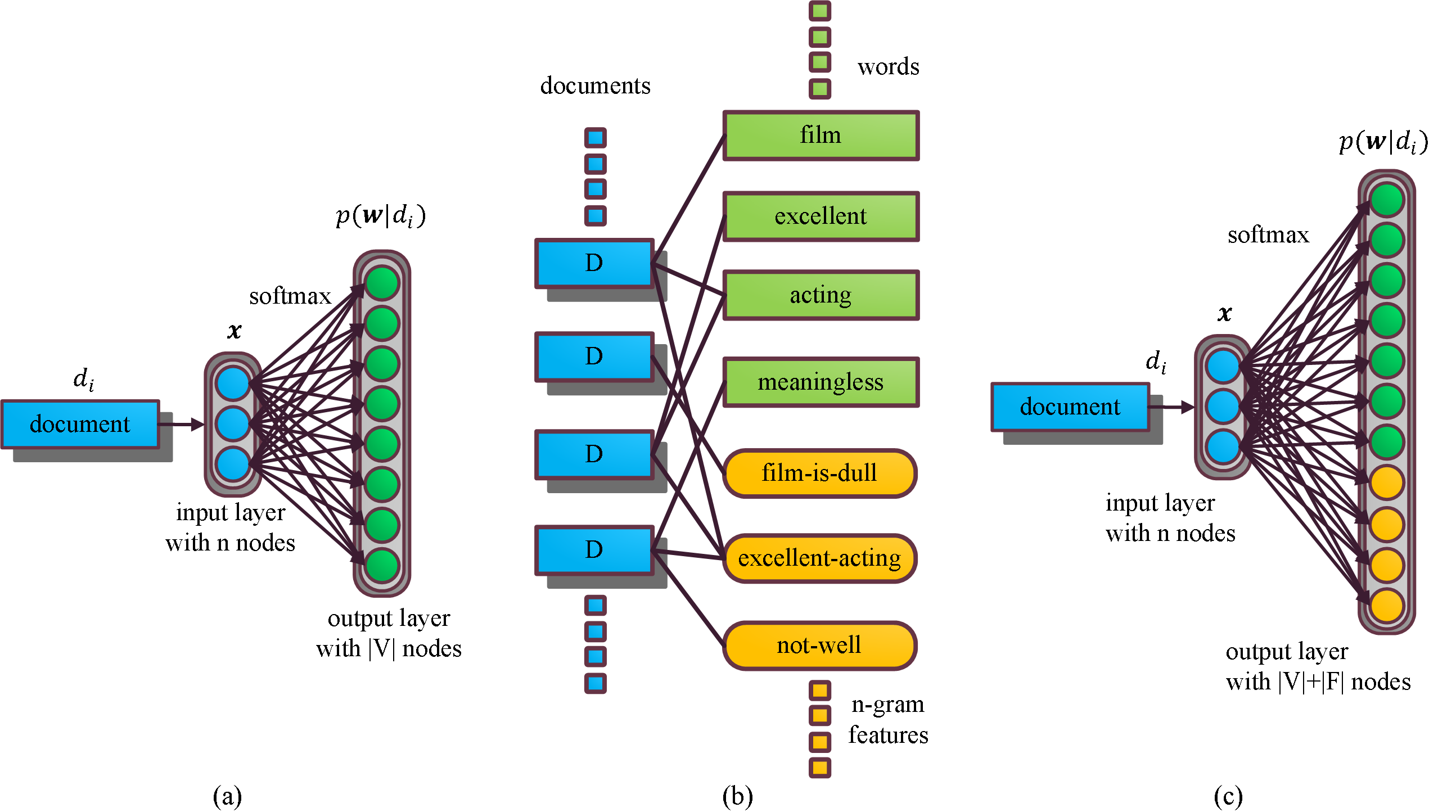}
\end{center}
\label{figure-combine}
\vspace{-1.5em}
\caption{(a) basic DV-ngram model. (b) illustration of n-gram features. (c) DV-ngram model.}
\end{figure}

In order to compute this probability, a simple neural network is built with a softmax output layer(as depicted in Figure~\ref{figure-combine}-a). The input layer of this network has $n$ nodes which represent the document vector, denoted by $x$. The output layer has $|V|$ (vocabulary size) nodes and the $k^{th}$ node represents the probability that the $k^{th}$ word belongs to this document. This probability can be written as
\begin{equation}\label{math-pez}
  \log p\left( w_{i,j} | d_{i} \right) = \frac{e^{y_{w_{i,j}}}}{Z}
\end{equation}
where $y_{w_{i,j}}$ is the unnormalized log-probability for each target word $w_{i,j}$, which can be calculated using $\mathbf{y}=\mathbf{b}+W\mathbf{x}$. $W$ and $\mathbf{b}$ are the network’s weights and biases. $Z$ in equation~\ref{math-pez} denotes the normalized factor which basically sums up all possible $e^{y_{w_{i,j}}}$

In our model, Stochastic Gradient Descent (SGD) \citep{williams1986learning} is used in all of our experiments for learning.

\subsection{Improved Model for Modeling Word Order}

Word order is often essential for understanding documents. For example, the following two texts have exact the same words but express totally different meanings due to their different word order: ``Despite dull acting, this film is excellent'', ``Despite excellent acting, this film is dull''. In order to model word order, distributed representation of documents is learned by predicting not only its belonging words but also word sequences. For simplicity, n-gram is directly used as word sequence features, which is illustrated by ``film-is-dull'', ``excellent-acting'' and ``not-well'' as shown in Figure~\ref{figure-combine}-b. More sophisticated word sequences selecting methods may be investigated in the future.

In practice, each word sequence is treated as a special token and is directly appended to each document. The output layer of the above neural network is also expanded as shown in Figure~\ref{figure-combine}-c. Thus, documents that contain semantically similar word sequences also tend to be closer to each other in vector space.

As shown later in our experiments, much better performance can be obtained by this improved model.

\subsection{Learning Acceleration}
In practice, since the size of vocabulary $V$ and feature set $F$ can be very large, our model needs to compute the output values of $|V|+|F|$ nodes in output layer, which results in computation inefficiency. Negative sampling technique \citep{Mikolov2013NIPS} is used to accelerate the training process. Negative sampling is especially efficient and simple, it only calculates the values of $K$ nodes ($K$ is a small constant) compared to standard softmax which calculates $|V|+|F|$ nodes in each training step. More precisely, negative sampling basically calculates equation~\ref{math-pwd} as
\begin{equation}\label{math-neg}
  \sum_{i} \sum_{j} \left[ f\left( x_{w_{i,j}}^\top x_{d_{i}} \right) + \sum_{k=1}^{K} f\left( -x_{w_{\text{random}}}^\top x_{d_{i}} \right) \right]
\end{equation}
where $x_{w_{i,j}}$ represents the vector of $j^{th}$ word/feature from $i^{th}$ document. $x_{d_{i}}$ represents the vector of $i^{th}$ document. $w_{\text{random}}$ represents the vector of word randomly sampled from the vocabulary based on words frequency. $K$ is the negative sampling size and $f$ is sigmoid function.

In summary, in order to get desired document vector, DV-ngram first randomly initialize each document vectors. Then stochastic gradient descent is used to maximize equation~\ref{math-neg} to get desired document vectors. The document vectors are eventually sent to a logistic regression classifier for sentiment classification. Note that DV-ngram use no labeled information thus is unsupervised. As shown in our experiments, additional unlabeled data can be use to improve model's performance.

\section{Experiments}
\subsection{Dataset and Experimental Setup}
Our model is benchmarked on well-studied IMDB sentiment classification dataset \citep{Maas2011ACL}. This dataset contains 100,000 movie reviews, of which 25,000 are positives, 25,000 are negatives and the rest 50,000 are unlabeled. Average document length of this dataset is 231 words. Accuracy is used to measure the performance of sentiment classification.

For comparison with other published results, we use the default train/test split for IMDB dataset. Since development data are not provided by two datasets, we refer the previous method of \citet{mesnil2014ensemble}, i.e. 20\% of training data are selected as development data to validate hyper-parameters and experiment settings, optimal results are shown in Table~\ref{table-param}.

\begin{table}[h]
\caption{Optimal hyper-parameters and experiment settings}
\label{table-param}
\begin{center}
\begin{tabular}{c c c c c}
\hline
Vector size & Learning rate & Mini-batch & Iteration & Negative sampling size \\
\hline
500 & 0.25 & 100 & 10 & 5 \\

\end{tabular}
\end{center}
\end{table}

Document vectors and parameters of neural network are randomly initialized with values uniformly distributed in the range of [-0.001, +0.001]. We use logistic regression classifier in LIBLINEAR package \citep{Fan2008JournalofMachineLearningResearch} \footnote{Available at \url{http://www.csie.ntu.edu.tw/~cjlin/liblinear/}} as the sentiment classifier.

In order to reduce the effect of random factors, training and testing were done for five times and the average of all the runs was obtained.

The experiments can be reproduced using our DV-ngram package, which can be found at \url{https://github.com/libofang/DV-ngram}.

\subsection{Comparison with Bag-of-ngram Baselines}
Our model is first evaluated by comparing with traditional bag-of-ngram baselines since they both use n-gram as feature. The biggest difference of these two kinds of methods is the way of representing documents. Bag-of-ngram methods use one-hot representation which loses semantics in some extent. DV-ngram is superior for modeling semantics since it represents documents by compact low dimension continuous-value vectors.

\begin{table}[h]
\caption{Comparison of DV-ngram with bag-of-ngram baseline. }
\label{table-baseline}
\begin{center}
\begin{tabular}{l l l l}
\hline
\textbf{Model} & \textbf{Unigram} & \textbf{Bigram} & \textbf{Trigram} \\
\hline
bag-of-ngram & 86.95 & 89.16 & 89.00 \\
DV-ngram (our model) & 89.12 & 90.63 & 91.75 \\
DV-ngram+Unlab’d (our model) & \textbf{89.60} & \textbf{91.27} & \textbf{92.14} \\

\end{tabular}
\end{center}
\end{table}

As shown in Table~\ref{table-baseline}, DV-ngram with different n-grams consistently outperforms corresponding bag-of-ngram methods. This results also suggests that the performance of DV-ngram can be further improved by adding more unlabeled sentiment related documents. Note that some other models are inherently unable to make use of this additional data such as the bag-of-ngram methods in this table. The best performance is achieved by DV-tri, for simplicity, we will report only the result of DV-tri in following experiments.

\subsection{Comparison with Other Models}

DV-ngram is compared with both traditional bag-of-ngram based models and deep learning models. Any type of model or feature combination is not considered here for comparison fairness, combination will be discussed later. Additional unlabeled documents are used by Maas, PV and DV-ngram when learning document vectors but not used by other methods since they are task specified.

As shown in Table~\ref{table-compare}, DV-ngram greatly outperforms most of other deep learning models. Especially, DV-tri outperforms PV 3.41 percent in terms of accuracy. This result shows that the prediction of word sequences is important for document embeddings. Note that even the simplest DV-uni (use words alone with no n-gram feature) outperforms PV 0.87 percent in terms of accuracy. This result supports our claim in Section~\ref{section-1} that the way PV handles context information may not suitable for sentiment analysis of movie reviews.

Among all other models, NBSVM is the most robust model for this dataset. NBSVM basically use labeled information to weight each words. Even though DV-ngram use no labeled information, it still outperforms NBSVM and achieves the new single model state-of-the-art results on IMDB dataset.

\subsection{Feature Combination}

\begin{table}[t]
\caption{Comparison of DV-ngram with other models. \tablefootnote{Result of DCNN is reported by \citet{Iyyer2015ACL}. Results of RNN-LM and PV are reported by \citet{mesnil2014ensemble} }}
\label{table-compare}
\begin{center}
\begin{tabular}{l l}
\hline
\textbf{Bag-of-ngram based models} & \textbf{Accuracy} \\
\hline
LDA \citep{Maas2011ACL}          & 67.42 \\
LSA \citep{Maas2011ACL}       & 83.96 \\
MNB-bi \citep{Wang2012ACL}    & 86.59 \\
NBSVM-bi \citep{Wang2012ACL}  & 91.22 \\
NBSVM-tri \citep{mesnil2014ensemble} & \textbf{91.87} \\
 & \\
\hline
\textbf{Deep learning models} & \textbf{Accuracy} \\
\hline
RNN-LM  \citep{mikolov2012statistical} & 86.60          \\
WRRBM \citep{Dahl2012ICML}                                  & 87.42           \\
DCNN \citep{kalchbrenner2014convolutional} & 89.4          \\
DAN \citep{Iyyer2015ACL}  & 89.4          \\
seq-CNN \citep{Johnson2015NAACL}    & 91.61 \\
DV-tri (our model)                                          & \textbf{91.75} \\
\hline
Maas \citep{Maas2011ACL}                            & 87.99    \\
PV \citep{Le2014ICML}   &   88.73           \\
DV-tri+Unlab'd (our model) & \textbf{92.14} \\

\end{tabular}
\end{center}
\end{table}

In practice, more sophisticated supervised features such as Naive Bayes weigted bag-of-ngram vectors (NB-BO-ngram) \citep{Wang2012ACL} can be used to improve performance of classification. Previous state-of-the-art results obtained by feature combination is achieved by an ensemble model named seq2-CNN \citep{Johnson2015NAACL}. The seq2-CNN model integrates three kind of vectors including NB-BO-ngram in a parallel convolutional neural network. For our model, we directly concatenate the learned document vectors with NB-BO-ngram for classification. As shown in table~\ref{table-combine}, when integrated with NB-BO-ngram, our model achieves new state-of-the-art result among feature combination models.

\begin{table}[h]
\caption{Different feature combination results.}
\label{table-combine}
\begin{center}
\begin{tabular}{l l l}
\hline
\textbf{Model} & \textbf{Alone} & \textbf{+NB-BO-tri}\\
\hline
seq2-CNN \citep{Johnson2015NAACL}                & 91.96  & 92.33 \\
DV-tri (our model)                          & 91.75  & 92.74\\
DV-tri+Unlab'd (our model)                  & \textbf{92.14}  & \textbf{92.91}\\
\end{tabular}
\end{center}
\end{table}

\subsection{Model Ensemble}

Recently, a new ensemble model \citep{mesnil2014ensemble} is proposed, which achieves the new state-of-the-art result for ensemble models on IMDB dataset. Optimal weights are obtained by grid search for each sub-model. In our experiment, we find the weights for different models are almost the same. For simplicity, we directly combine our model with others without weighting.

As shown in Table~\ref{table-ensemble}, the previous best performance is obtained by combining PV, RNN-LM and NBSVM (NBSVM with trigram). Without much surprise, a new state-of-the-art result is obtained by replacing PV to our model. Note that combining with or without RNN-LM do not affect results much. One reason for this may be that RNN-LM becomes burdensome when combined with more robust model since RNN-LM alone only achieves 86.6 percent in terms of accuracy.

\begin{table}[h]
\caption{Different model ensemble results. R: RNN-LM. N: NBSVM.}
\label{table-ensemble}
\begin{center}
\begin{tabular}{l l l l l}
\hline
\textbf{Model} & \textbf{Alone} & \textbf{+R} & \textbf{+N} & \textbf{+R+N} \\
\hline
PV \citep{mesnil2014ensemble}   & 88.73	& 90.40	& 92.39	& 92.57 \\
DV-tri (our model)                          & 91.75 & 92.10 & 92.81 & 92.89 \\
DV-tri+Unlab'd (our model)                  & \textbf{92.14} & \textbf{92.31} & \textbf{93.00} & \textbf{93.05} \\
\end{tabular}
\end{center}
\end{table}

\section{Conclusion}

A new method for learning document embeddings has been proposed for sentiment analysis of movie reviews, which is based on recently proposed Paragraph Vector. Instead of learning both document vectors and word vectors simultaneously by predicting the target word, our model learns document vectors alone by predicting both their belonging words and n-gram features. In this way, the expressive power of document vectors is kept. Experimental results show that the proposed model outperforms PV due to this reason.

Furthermore, comparing with traditional bag-of-ngram models, our model can represent the semantics which is important for sentiment analysis. Our model is also compared with other deep learning and bag-of-ngram based models and achieves the state-of-the-art results on IMDB dataset. We also show that the performance of our model can be further improved by adding unlabeled data.

Finally, when combined with NBSVM and RNN-LM, our model achieves state-of-the-art result among all other ensemble models.

The source code of our model will be published together with this paper. We hope this could allow researchers to reproduce our experiments easily for further improvements and applications to other tasks.

\section{Acknowledgements}

This work is supported by National Natural Science Foundation of China (61472428, 61003204), Tencent company, the Fundamental Research Funds for the Central Universities, the Research Funds of Renmin University of China No. 14XNLQ06.

\bibliography{L}

\begin{thebibliography}{21}
\providecommand{\natexlab}[1]{#1}
\providecommand{\url}[1]{\texttt{#1}}
\expandafter\ifx\csname urlstyle\endcsname\relax
  \providecommand{\doi}[1]{doi: #1}\else
  \providecommand{\doi}{doi: \begingroup \urlstyle{rm}\Url}\fi

\bibitem[Bengio et~al.(2003)Bengio, Ducharme, Vincent, and
  Janvin]{bengio2003neural}
Bengio, Yoshua, Ducharme, R{\'e}jean, Vincent, Pascal, and Janvin, Christian.
\newblock A neural probabilistic language model.
\newblock \emph{The Journal of Machine Learning Research}, 3:\penalty0
  1137--1155, 2003.

\bibitem[Collobert \& Weston(2008)Collobert and Weston]{collobert2008unified}
Collobert, Ronan and Weston, Jason.
\newblock A unified architecture for natural language processing: Deep neural
  networks with multitask learning.
\newblock In \emph{Proceedings of the 25th international conference on Machine
  learning}, pp.\  160--167. ACM, 2008.

\bibitem[Collobert et~al.(2011)Collobert, Weston, Bottou, Karlen, Kavukcuoglu,
  and Kuksa]{collobert2011natural}
Collobert, Ronan, Weston, Jason, Bottou, L{\'e}on, Karlen, Michael,
  Kavukcuoglu, Koray, and Kuksa, Pavel.
\newblock Natural language processing (almost) from scratch.
\newblock \emph{The Journal of Machine Learning Research}, 12:\penalty0
  2493--2537, 2011.

\bibitem[Dahl et~al.(2012)Dahl, Adams, and Larochelle]{Dahl2012ICML}
Dahl, George~E., Adams, Ryan~Prescott, and Larochelle, Hugo.
\newblock Training restricted boltzmann machines on word observations.
\newblock In \emph{ICML}, 2012.

\bibitem[Fan et~al.(2008)Fan, Chang, Hsieh, Wang, and
  Lin]{Fan2008JournalofMachineLearningResearch}
Fan, Rong-En, Chang, Kai-Wei, Hsieh, Cho-Jui, Wang, Xiang-Rui, and Lin,
  Chih-Jen.
\newblock Liblinear: A library for large linear classification.
\newblock \emph{Journal of Machine Learning Research}, 9:\penalty0 1871--1874,
  2008.

\bibitem[Iyyer et~al.(2015)Iyyer, Manjunatha, Boyd-Graber, and
  III]{Iyyer2015ACL}
Iyyer, Mohit, Manjunatha, Varun, Boyd-Graber, Jordan~L., and III, Hal~Daumé.
\newblock Deep unordered composition rivals syntactic methods for text
  classification.
\newblock In \emph{ACL}, 2015.

\bibitem[Johnson \& Zhang(2015)Johnson and Zhang]{Johnson2015NAACL}
Johnson, Rie and Zhang, Tong.
\newblock Effective use of word order for text categorization with
  convolutional neural networks.
\newblock In \emph{NAACL}, 2015.

\bibitem[Kalchbrenner et~al.(2014)Kalchbrenner, Grefenstette, and
  Blunsom]{kalchbrenner2014convolutional}
Kalchbrenner, Nal, Grefenstette, Edward, and Blunsom, Phil.
\newblock A convolutional neural network for modelling sentences.
\newblock In \emph{ACL}, 2014.

\bibitem[Kim(2014)]{Kim2014EMNLP}
Kim, Yoon.
\newblock Convolutional neural networks for sentence classification.
\newblock In \emph{EMNLP}, 2014.

\bibitem[Kiros et~al.(2015)Kiros, Zhu, Salakhutdinov, Zemel, Torralba, Urtasun,
  and Fidler]{kiros2015skip}
Kiros, Ryan, Zhu, Yukun, Salakhutdinov, Ruslan, Zemel, Richard~S, Torralba,
  Antonio, Urtasun, Raquel, and Fidler, Sanja.
\newblock Skip-thought vectors.
\newblock \emph{arXiv preprint arXiv:1506.06726}, 2015.

\bibitem[Le \& Mikolov(2014)Le and Mikolov]{Le2014ICML}
Le, Quoc~V. and Mikolov, Tomas.
\newblock Distributed representations of sentences and documents.
\newblock In \emph{ICML}, 2014.

\bibitem[Maas et~al.(2011)Maas, Daly, Pham, Huang, Ng, and Potts]{Maas2011ACL}
Maas, Andrew~L., Daly, Raymond~E., Pham, Peter~T., Huang, Dan, Ng, Andrew~Y.,
  and Potts, Christopher.
\newblock Learning word vectors for sentiment analysis.
\newblock In \emph{ACL}, 2011.

\bibitem[Mesnil et~al.(2014)Mesnil, Ranzato, Mikolov, and
  Bengio]{mesnil2014ensemble}
Mesnil, Gr{\'e}goire, Ranzato, Marc'Aurelio, Mikolov, Tomas, and Bengio,
  Yoshua.
\newblock Ensemble of generative and discriminative techniques for sentiment
  analysis of movie reviews.
\newblock \emph{arXiv preprint arXiv:1412.5335}, 2014.

\bibitem[Mikolov(2012)]{mikolov2012statistical}
Mikolov, Tom{\'a}{\v{s}}.
\newblock Statistical language models based on neural networks.
\newblock \emph{Presentation at Google, Mountain View, 2nd April}, 2012.

\bibitem[Mikolov et~al.(2013)Mikolov, Sutskever, Chen, Corrado, and
  Dean]{Mikolov2013NIPS}
Mikolov, Tomas, Sutskever, Ilya, Chen, Kai, Corrado, Gregory~S., and Dean,
  Jeffrey.
\newblock Distributed representations of words and phrases and their
  compositionality.
\newblock In \emph{NIPS}, 2013.

\bibitem[Schwenk(2007)]{schwenk2007continuous}
Schwenk, Holger.
\newblock Continuous space language models.
\newblock \emph{Computer Speech \& Language}, 21\penalty0 (3):\penalty0
  492--518, 2007.

\bibitem[Socher et~al.(2011)Socher, Lin, Ng, and Manning]{Socher2011ICML}
Socher, Richard, Lin, Cliff Chiung-Yu, Ng, Andrew~Y., and Manning,
  Christopher~D.
\newblock Parsing natural scenes and natural language with recursive neural
  networks.
\newblock In \emph{ICML}, 2011.

\bibitem[Socher et~al.(2013)Socher, Perelygin, Wu, Chuang, Manning, Ng, and
  Potts]{socher2013recursive}
Socher, Richard, Perelygin, Alex, Wu, Jean~Y, Chuang, Jason, Manning,
  Christopher~D, Ng, Andrew~Y, and Potts, Christopher.
\newblock Recursive deep models for semantic compositionality over a sentiment
  treebank.
\newblock In \emph{Proceedings of the conference on empirical methods in
  natural language processing (EMNLP)}, volume 1631, pp.\  1642. Citeseer,
  2013.

\bibitem[Srivastava et~al.(2013)Srivastava, Salakhutdinov, and
  Hinton]{Srivastava2013UAI}
Srivastava, Nitish, Salakhutdinov, Ruslan, and Hinton, Geoffrey~E.
\newblock Modeling documents with deep boltzmann machines.
\newblock In \emph{UAI}, 2013.

\bibitem[Wang \& Manning(2012)Wang and Manning]{Wang2012ACL}
Wang, Sida~I. and Manning, Christopher~D.
\newblock Baselines and bigrams: Simple, good sentiment and topic
  classification.
\newblock In \emph{ACL}, 2012.

\bibitem[Williams \& Hinton(1986)Williams and Hinton]{williams1986learning}
Williams, DE Rumelhart GE Hinton~RJ and Hinton, GE.
\newblock Learning representations by back-propagating errors.
\newblock \emph{Nature}, pp.\  323--533, 1986.

\end{thebibliography}
\bibliographystyle{L}

\end{document}